\UseRawInputEncoding
\documentclass[runningheads]{llncs}
\usepackage{graphicx}
\usepackage{blindtext} 
\usepackage{siunitx}
\usepackage{amssymb}
\usepackage{amsmath}

\begin{document}
\title{Cascaded Cross-Attention Networks for Data-Efficient Whole-Slide Image Classification Using Transformers}
%
%
\author{Firas Khader\inst{1} \and
Jakob Nikolas Kather\inst{2} \and
Tianyu Han\inst{3} \and
Sven Nebelung\inst{1} \and
Christiane Kuhl\inst{1} \and
Johannes Stegmaier\inst{4} \and
Daniel Truhn\inst{1}
}

%
\authorrunning{F. Khader et al.}
%
\institute{
 Department of Diagnostic and Interventional Radiology, University Hospital Aachen, Aachen, Germany \and
Physics of Molecular Imaging Systems, Experimental Molecular Imaging, RWTH Aachen 
Else Kroener Fresenius Center for Digital Health, Medical Faculty Carl Gustav Carus, 
Institute of Imaging and Computer Vision, RWTH Aachen University, Aachen, Germany
}

\maketitle              
\begin{abstract}
Whole-Slide Imaging allows for the capturing and digitization of high-resolution images of histological specimen. An automated analysis of such images using deep learning models is therefore of high demand. The transformer architecture has been proposed as a possible candidate for effectively leveraging the high-resolution information. Here, the whole-slide image is partitioned into smaller image patches and feature tokens are extracted from these image patches. However, while the conventional transformer allows for a simultaneous processing of a large set of input tokens, the computational demand scales quadratically with the number of input tokens and thus quadratically with the number of image patches. To address this problem we propose a novel cascaded cross-attention network (CCAN) based on the cross-attention mechanism that scales linearly with the number of extracted patches. Our experiments demonstrate that this architecture is at least on-par with and even outperforms other attention-based state-of-the-art methods on two public datasets: On the use-case of lung cancer (TCGA NSCLC) our model reaches a mean area under the receiver operating characteristic (AUC) of 0.970 $\pm$ 0.008 and on renal cancer (TCGA RCC) reaches a mean AUC of 0.985 $\pm$ 0.004. Furthermore, we show that our proposed model is efficient in low-data regimes, making it a promising approach for analyzing whole-slide images in resource-limited settings. To foster research in this direction, we make our code publicly available on GitHub: XXX.

\end{abstract}

\section{Introduction}
Computational pathology is an emerging interdisciplinary field that synergizes knowledge from pathology and computer science to aid the diagnoses and treatment of diseases \cite{cui_artificial_2021}. With the number of digitized pathology images increasing over the years, novel machine learning algorithms are required to analyse the images in a timely manner \cite{marini_unleashing_2022}. Deep learning-based methods have demonstrated a promising potential in handling image classification tasks in computational pathology \cite{kather_deep_2019}. A prevalent procedure for employing deep learning techniques to the analysis of whole-slide images (WSI) is multiple instance learning (MIL), where a training sample comprises a set of instances and a label for the whole set. The set of instances is commonly chosen to be the set of feature tokens pertaining to one WSI. These feature tokens are constructed by an initial extraction of patches contained in a single WSI and a subsequent utilization of pre-trained feature extractors to capture meaningful feature representations of these patches. This is followed by a feature aggregation, in which the feature representations of each WSI are combined to arrive at a final prediction. Naive approaches for aggregating the features consist of mean or max pooling but only achieve limited performances \cite{wang_revisiting_2018,kanavati_weakly-supervised_2020}. This has given rise to more sophisticated feature aggregation techniques specifically tuned to the field of computational pathology \cite{tomita_attention-based_2019,hashimoto_multi-scale_2020,lu_data-efficient_2021,ilse_attention-based_2018}. These are largely based on an attention operation, in which trainable parameters are used to compute the contribution of each instance to the final prediction \cite{ilse_attention-based_2018,tomita_attention-based_2019}. Other approaches employ graph neural networks to the task of MSI classification \cite{tu_multiple_2019,konda_graph_2020} or frameworks that use clustering approaches on the patches of WSI \cite{sharma_cluster--conquer_2021}. 
\newline\newline
More recently, the transformer architecture \cite{vaswani_attention_2017} has been introduced to the field of computational pathology \cite{shao_transmil_2021}. Transformer architectures have demonstrated state-of-the-art performance in the fields of natural language processing and computer vision \cite{dosovitskiy_image_2020}. The transformer model can be seen as an input agnostic method, that leverages the self-attention mechanism in order to aggregate its input tokens. However, one notable short-coming is the quadratic scaling of the self-attention mechanism with respect to the sequence length \cite{jaegle_perceiver_2021}. This becomes particularly problematic in the context of computational pathology, where whole-slide images are often several gigapixels large \cite{marini_unleashing_2022}, therefore resulting in thousands of image patches that have to be input into the transformer model. As a result, bigger GPUs are necessary to store the attention matrix in memory and the increasing number of compute operations limits training and inference speeds as well as model complexity. To overcome this problem, we present a novel neural network architecture that is based on the cross-attention mechanism \cite{jaegle_perceiver_2021,vaswani_attention_2017} and show that it is capable of efficiently aggregating feature tokens of whole-slide images while demonstrating state-of-the-art performance on two publicly available datasets. Furthermore, we show how attention maps can be extracted for our network and therefore allow for an increased interpretability. Finally, we demonstrate that our model is superior to the previous state of the art in situations where only a small number of training samples is available.

\section{Materials \& Methods}
\begin{figure}[h]
\centering
\includegraphics[width=0.85\textwidth]{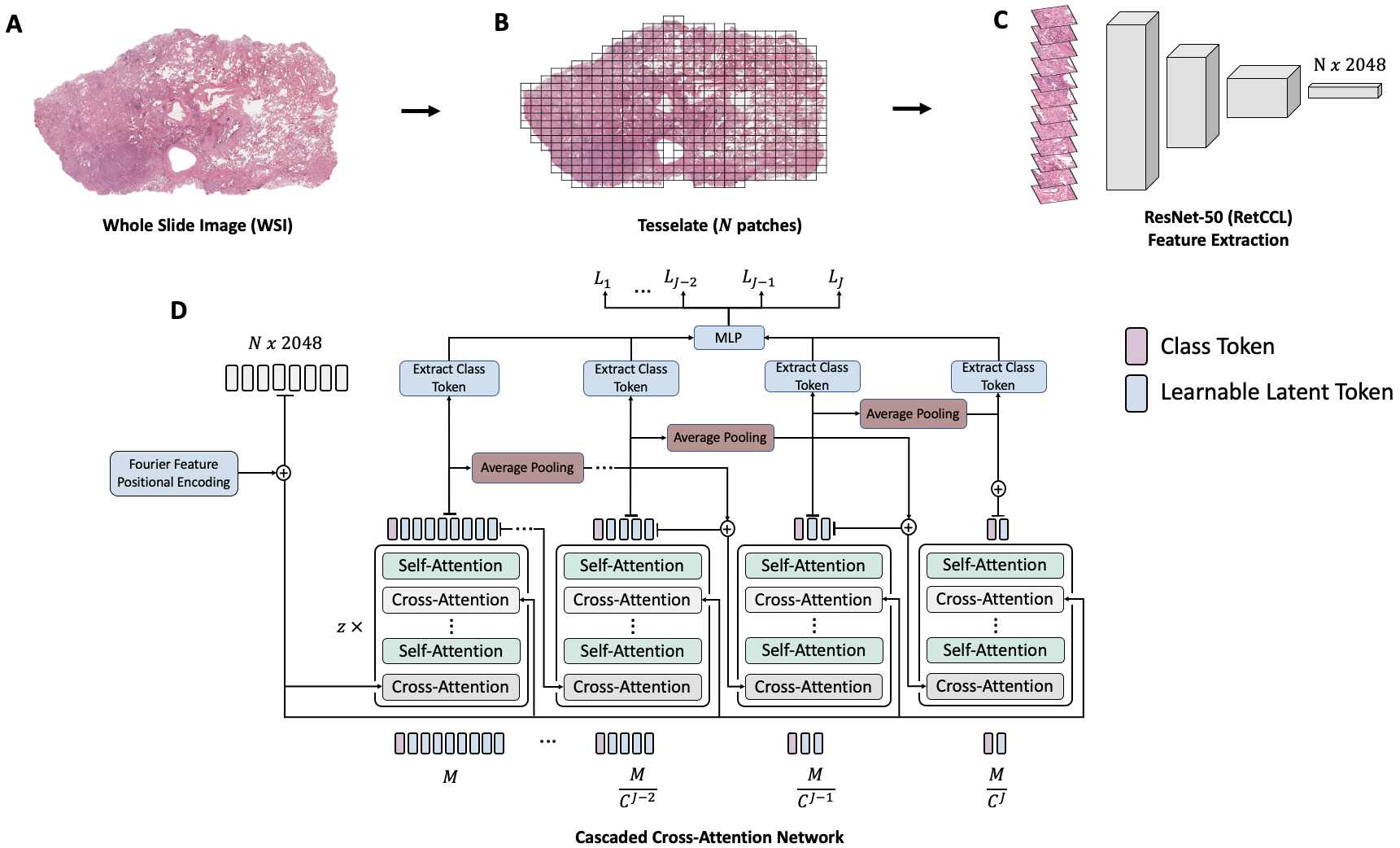}
\caption{Model architecture and pre-processing steps. In a first step, we extract a set of $N$ patches from each WSI and subsequently derive feature tokens based on a pretrained ResNet-50 model \cite{wang_retccl_2023} The resulting feature tokens are used as input to our model, where the cross attention mechanism is used to distil the information at each stage into a compressed representation that allows for efficient attention computation.} \label{model_architecture}
\end{figure}

\subsection{Dataset}
We demonstrate the performance of our model on two publicly available datasets: (1) First, the TCGA-NSCLC (non-small-cell lung cancer) dataset, which is constructed by merging the lung cancer datasets TCGA-LUSC and TCGA-LUAD. This results in a total of n=1,042 whole-slide images, of which 530 are labeled as LUAD (Lung adenocarcinoma) and 512 are labeled as LUSC (Lung squamous cell carcinoma). We proceed by splitting the dataset into training (60\%, n=627), validation (15\%, n=145) and testing sets (25\%, n=270) using a 4-fold cross validation scheme, while ensuring that images of the same patient only occur in the same set. (2) Additionally, we evaluate our model on the TCGA-RCC (renal cell carcinoma) dataset, which is composed of three other datasets: TCGA-KIRC, TCGA-KIRP and TCGA-KICH. This results in a dataset comprising n=937 whole-slide images, out of which 519 are labeled as KIRC (Kidney renal clear cell carcinoma), 297 labeled as KIRP (Kidney renal papillary cell carcinoma) and 121 labeled as KICH (Kidney Chromophobe). Similarly, we split the data into training (60\%, n=561), validation (15\%, n=141), and test sets (25\%, n=235) following a 4-fold cross validation scheme.

\subsection{Preprocessing}
Prior to inputting the WSIs into the neural network, various pre-processings steps are executed on each image: First, the whole-slide image is tesselated by extracting non-overlapping square patches with a \SI{256}{\micro\metre} edge length. These are then further resized to $256 \times 256$ pixels. Patches that possess a predominantly white color are removed by filtering out patches with a mean grayscale pixel value greater than 224. Furthermore, blurry images are excluded by assessing the fraction of pixels that belong to an edge within the patch using a Canny Edge Detector \cite{canny_computational_1986} and rejecting patches in which this fraction is below 2\%. This results in a mean number of 3,091 patches per WSI (min: 38; max: 11,039) for TCGA-NSCLC, and a mean number of 3,400 patches per WSI (min: 90; max: 10,037) for TCGA-RCC. In a second step, feature tokens are obtained from each patch using the RetCCL \cite{wang_retccl_2023} feature extractor, which is based on a ResNet-50 architecture \cite{he_deep_2016}. The output feature tokens have a dimension of 2,048.

\subsection{Architecture}
One of the key limitations of the conventional transformer model when applied to pathology images is the quadratic scaling of the number of compute operations and the GPU memory footprint with respect to the number of input tokens \cite{vaswani_attention_2017}. In order to overcome this problem of quadratic scaling, we base our CCAN architecture (see Figure \ref{model_architecture}) on the cross-attention mechanism proposed by Jaegle et al. \cite{jaegle_perceiver_2021}. The idea is to distil the information contained in the feature tokens of each MSI into a smaller set of latent tokens. More precisely, let $N \in \mathbb{R}^{D_{f}}$, denote the number of feature tokens of dimension $D_{f}$ extracted from each patch of the WSI and let $M  \in \mathbb{R}^{D_{l}}$ denote a pre-defined set of learnable latent tokens with dimensionality $D_{l}$. To provide the neural network with information about the position of each patch in the WSI, we first concatenate to each feacture token a positional encoding based on Fourier features \cite{jaegle_perceiver_2021}:
\begin{equation}
    p = [\sin(f_i \pi \hat D), \cos(f_i \pi \hat D)]
\end{equation}
Here $f_i$ denotes the $i^{th}$ frequency term contained in a set of $I$ equidistant frequencies between 1 and $f_{max}$ and $\hat D$ denotes the location of the feature token, normalized between -1 and 1, thereby mapping the top left feature token in the WSI to -1 and the bottom right token to 1. We motivate the use of Fourier features for positional encoding over the use of learnable encodings because they enable scaling to arbitrary sequence lengths. This property is particularly useful when dealing with WSI, which often present in varying sizes. We proceed by computing a key $K_f \in \mathbb{R}^{N \times D_l}$ and value $V_f \in \mathbb{R}^{N \times D_l}$ vector for each of the $N$ input tokens, and a query $Q_l \in \mathbb{R}^{M \times D_l}$ vector for each of the $M$ latent tokens, such that each latent token can attend to all of the feature tokens of the input image:

\begin{equation}
    f_{\text{Cross-Attention}}(Q_l, K_f, V_f) = softmax\Bigl(\frac{Q_lK_f^T}{\sqrt{M}}\Bigr)V_f
\end{equation}
Provided that $M \ll N$, the network learns a compressed representation of the information contained in the large set of input tokens. In addition the number of computation steps reduces to $O(MN)$, compared to $O(N^2)$ as is the case when self-attention is applied directly to the input tokens. Thus, by choosing $M$ to be much smaller than $N$, the computational requirements for processing a whole-slide image can be significantly reduced.
The information in the $M$ latent tokens is then further processed through self-attention layers that scale with $O(M^2)$. The cross-attention and self-attention blocks are then repeated $Z$ times. 
\newline\newline
When training neural networks, it is often beneficial to gradually change the input dimension opposed to abruptly reducing it (in convolutional neural networks the kernel size and stride are chosen in a way that gradually reduces the feature map dimensions at each stage). Based on this intuition, we propose to use a multi-stage approach in which the set of $M$ tokens is further distilled by adding additional stages with a reduced set of $\frac{M}{C}$ latent tokens. $C$ denotes an arbitrary compression factor and is chosen to be 2 in our model. Similar to the previous stage, cross-attention is used to distil the information, followed by self-attention blocks to process the compressed representation. Subsequently, we add the output of stage 1 to the output of stage 2, thereby serving as a skip-connection which allows for improved gradient flow when the number of stages $J$ is chosen to be large \cite{he_deep_2016}. To overcome the dimensionality mismatch that occurs when adding the $M$ tokens of stage 1 to the $\frac{M}{C}$ tokens of stage 2, we aggregate the output tokens of stage 1 by means of an average pooling that is performed on each set of $C$ consecutive tokens. This multi-stage approach is repeated for $J$ stages, whereby in each stage the number of tokens of the previous stage is compressed by the factor $C$. 
\newline\newline
One limitation of the multi-stage approach and the repeated cross- and self attention blocks is that the computation of attention maps (using e.g. the attention rollout mechanism) becomes unintuitive. To overcome this limitation, we add a class token to each of the stages $j \in J$ (resuling in $\frac{M}{C^j}$ latent tokens and 1 class token), and feed it into a shared multi-layer perceptron, resulting in a prediction $p_j$ and loss term $L_j$ for each of the $J$ stages. Additionally, in each stage we add a final cross-attention layer that takes as input the original set of $N$ feature tokens, followed by a self-attention block. This allows us to visualize what regions in the WSI the class token is attending to at each stage. The final prediction is computed by averaging all the individual contributions $p_j$ of each stage. Similarly, we backpropagate and compute the total loss $L_{\text{total}}$ by summing over all individual loss terms: 
\begin{equation}
    L_{\text{total}} = \Sigma_{j=1}^{J} L_j
\end{equation}
In our experiments, we chose the binary cross entropy loss as our loss function. Furthermore, to prevent the neural networks from overfitting, we randomly dropout a fraction $p_{do}$ of the input feature tokens. All models are trained on an NVIDIA RTX A6000 GPU for a total of 100 epochs and were implemented using PyTorch 1.13.1. We chose the best model of each run in terms of the highest area under the receiver operating characteristic (AUC) reached on the validation set. Further details regarding the hyperparameters used can be found in Supplementary Table \ref{tab:hyperparams}. 

\section{Results}
\subsection{Baseline}
To assess the performance of our model, we compare its performance to TransMIL \cite{shao_transmil_2021}, a popular transformed-based method proposed for aggregating feature tokens in a MIL setting. TransMIL has demonstrated state-of-the-art performance on a number of datasets, outperforming other popular MIL techniques by a notable margin (see Supplementary Table \ref{tab:transmil_comparison} for a comparison to other methods). In essence, feature tokens are extracted from a set of patches pertaining to each WSI, and then fed through two transformer encoder layers. To handle the quadratic scaling of the self-attention mechanism with respect to the number of input tokens, they make use of Nyström attention \cite{xiong_nystromformer_2021} in each transformer layer. The low-rank matrix approximation thereby allows for a linear scaling of the attention mechanism. In addition, the authors of the architecture propose the use of a convolutional neural network-based pyramid position encoding generator to encode the positional information of each feature token.   

\subsection{WSI classification}
\begin{figure}[h]
\includegraphics[width=0.95\textwidth]{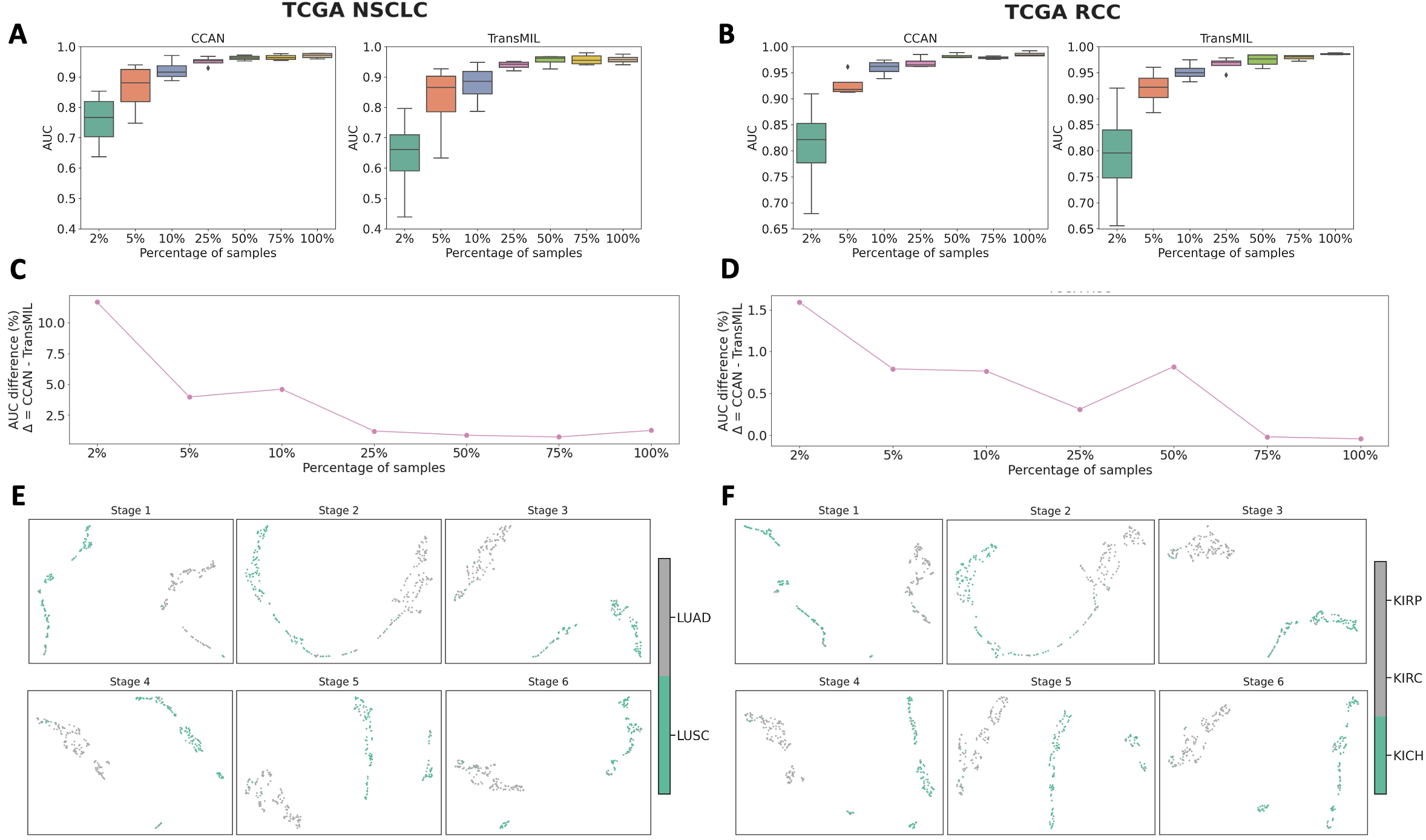}
\centering
\caption{Results of the CCAN and TransMIL models when trained on the TCGA NSCLC (A) and TCGA RCC (B) datasets. Boxplots show the results of the 4-fold cross validation for each fraction of the used training dataset. (C) and (D) thereby visualize the difference in the mean AUC between the two models, showing that CCAN is more data efficient when a small portion of training data is used. Finally, UMAP \cite{mcinnes_umap_2018} projections of the class token of CCAN are visualized at each stage for TCGA NSCLC (E) and TCGA RCC (F), demonstrating separable clusters for all stages that contribute to the final prediction of the model.} \label{results}
\end{figure}

We compare the performance of the models on two publicly available datasets, comprising cancer subtype classification tasks, TCGA NSCLC and TCGA RCC (https://portal.gdc.cancer.gov/projects/). For the binary classification task (i.e., LUSC vs. LUAD in TCGA NSCLC), we use the AUC, to assess the model performance. For the multiclass classification problem (i.e., KICH vs. KIRC vs. KIRP in TCGA RCC), we assess the performance by computing the (macro-averaged) one vs. rest AUC. To guarantee a fair comparison, we train all models on the exact same data splits of the 4-fold cross validation. For TCGA NSCLC, we find that our model outperforms the TransMIL baseline (AUC: 0.970 $\pm$ 0.008 [standard deviation -
SD] vs. 0.957 $\pm$ 0.013 [SD]) thereby demonstrating a strong performance. Similarly, we find that on the TCGA RCC dataset our model is on par with the state-of-the-art results set by TransMIL (both reach an AUC of 0.985 [SD CCAN: 0.004, SD TransMIL: 0.002], see Figure \ref{results} for a more detailed comparison). 

\subsection{Data Efficiency}
A key obstacle in the field of medical artificial intelligence is the restricted access to large datasets. Therefore, neural network architectures need to be developed that can attain a high performance, even under conditions of limited available data. To simulate the performance in limited data settings, we train our model multiple times and restrict the amount of data seen during training to 2\%, 5\%, 10\%, 25\%, 50\%, 75\% and 100\%. For comparison, we similarly train the baseline method on the same training data. For the TCGA NSCLC dataset, we find that with only 2\% of data, our model (AUC: 0.756 $\pm$ 0.095 [SD]) outperforms the previous state-of-the-art method by 12\% (AUC: 0.639 $\pm$ 0.149 [SD]). Similar results were found for other dataset sizes (see Figure \ref{results}). Accordingly, for the TCGA RCC dataset, our model (AUC: 0.808 $\pm$ 0.096 [SD]) outperforms the baseline (AUC: 0.792 $\pm$ 0.109 [SD]) at only 2\% of training data by 1.6\%. Again, this trend of a superior performance can be seen for all other percentages of the training data.

\subsection{Explainability}
\begin{figure}[h]
\centering
\includegraphics[width=0.73\textwidth]{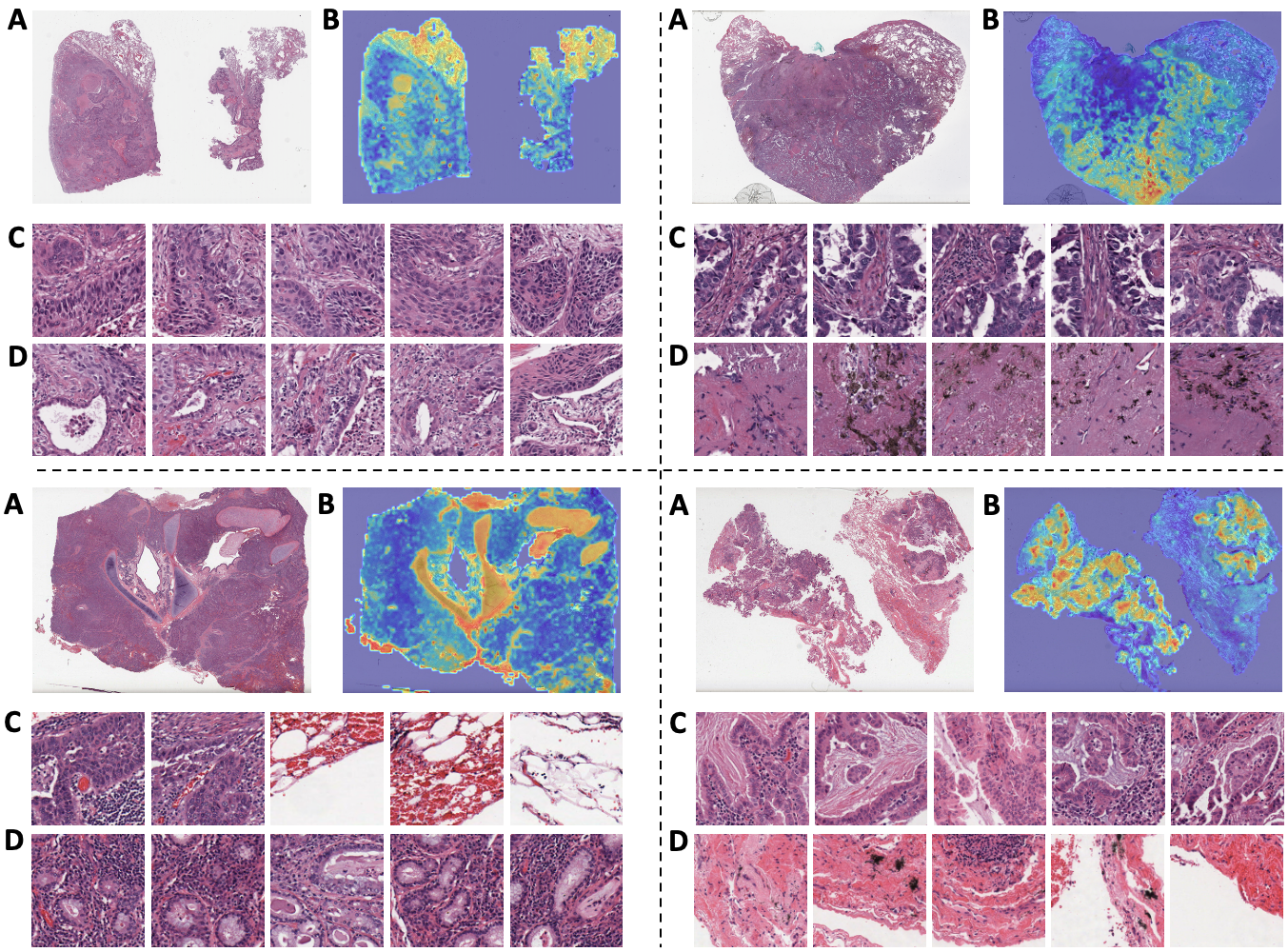}
\caption{Attention visualization for four different WSI contained in the TCGA NSCLC dataset. (A) shows the original WSI and (B) displays the aggregated attention map over all stages. Additionally, the top-5 patches with the lowest attention value (C) and top-5 patches with highest attention value (D) are visualized.} \label{attention_vis}
\end{figure}

In order to gain a deeper understanding of the inner workings of the neural network, attention maps were extracted. Since the class tokens are used as input in the final MLP and the output thereof is used to arrive at the final prediction, we propose to compute the attention map of the class token with respect to the input patches. More precisely, in each stage we perform attention-rollout \cite{abnar_quantifying_2020} using the last self-attention layers and the last cross-attention layer. We then use computed attention values to visualize the attention map of each class token, thereby providing insights into the decision making process of the model at each stage (see Supplementary Figure \ref{attention_vis_stages}). To arrive at a single attention map for the whole model, we take the mean attention over all stages. We find that these attention maps largely coincide with regions containing the tumor (see Figure \ref{attention_vis}). Additionally, we display regions of the model with low and high attention values. 
\section{Conclusion}
In this work we developed and presented a novel neural network architecture, based on the cross-attention mechanism, that is capable of aggregating a large set of input feature vectors previously extracted from whole-slide images. We demonstrate that this method outperforms the previous state-of-the-art methods on two publicly available datasets. In particular we show that our model architecture is more data efficient when training data is limited. Furthermore, we provide insights into the models decision-making process by showing how attention maps of each stage in the model can be aggregated. This allows for an interpretable visualization of which regions of the image the neural networks looks at to arrive at a final prediction. The model design lends itself to be extended to support multimodal inputs such as WSI in combination with genomics, which we will explore in future work.

\section{Acknowledgement}
The results published here are in whole or part based upon data generated by the TCGA Research Network: https://www.cancer.gov/tcga.

%
%
%
%
\newpage
\bibliographystyle{splncs04}
\bibliography{references_cleaned}

%
%
%
%

\newpage
\section{Supplemental Material}
\renewcommand{\thefigure}{S\arabic{figure}}
\renewcommand{\thetable}{S\arabic{table}}
\setcounter{figure}{0}    
\setcounter{table}{0}    

\begin{figure}
\includegraphics[width=\textwidth]{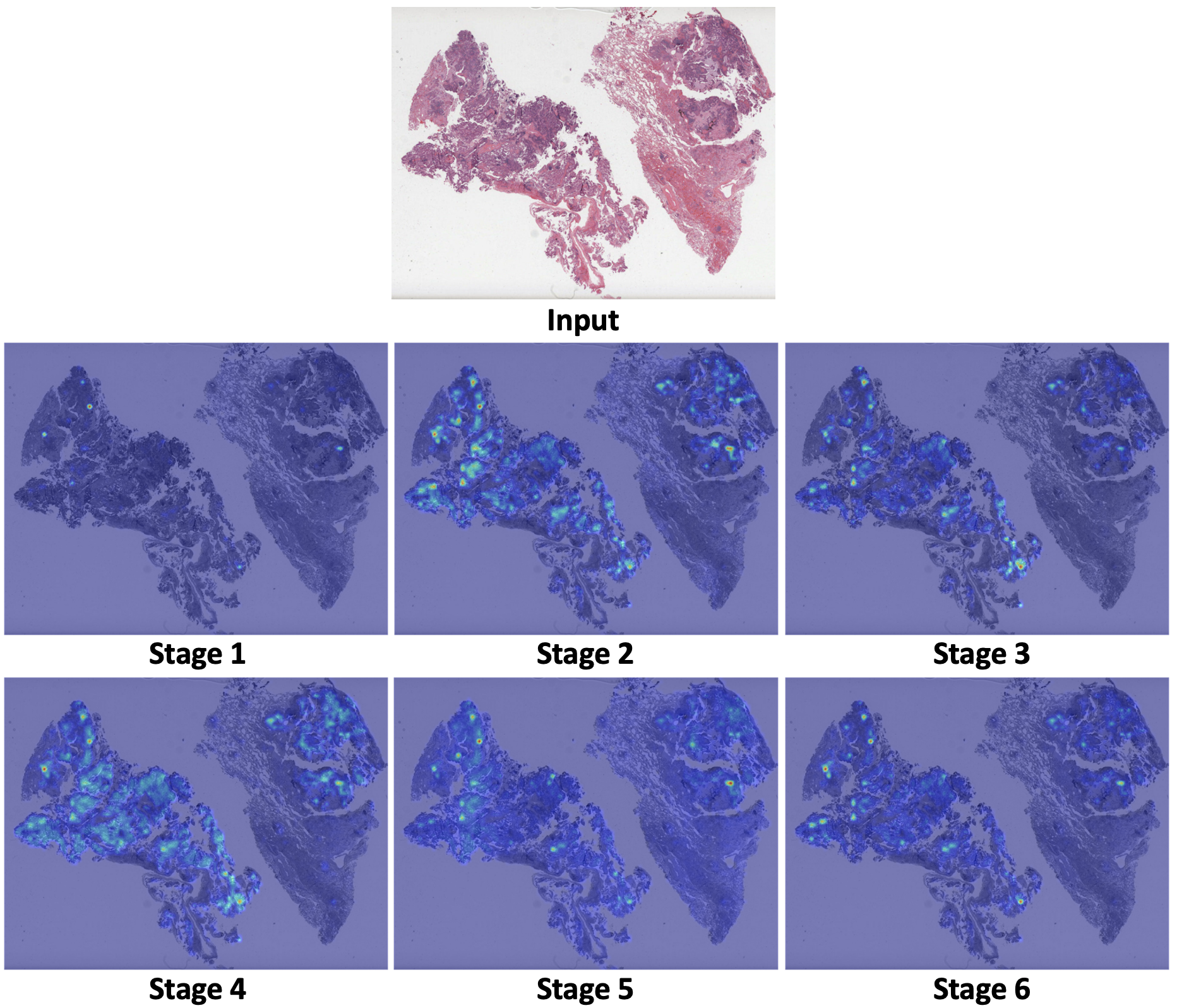}
\caption{Individual attention maps for each stage. The multi-stage approach of CCAN allows to display the attention of the class token with respect to the input image at each stage. We find that at each stage, the neural network focuses on different parts of the image.} \label{attention_vis_stages}
\end{figure}

\setlength{\tabcolsep}{0.5em}
{\renewcommand{\arraystretch}{1.4}
\begin{table}[]
\centering
\caption{Comparison of TransMIL to other methods in terms of area under the receiver operating characteristic (taken from \cite{shao_transmil_2021}). For completeness, we also provide the results of TransMIL as well as CCAN on our data splits.}
\label{tab:transmil_comparison}
\begin{tabular}{l|c|c|}
\cline{2-3}
                                            & \textbf{TCGA NSCLC} & \textbf{TCGA RCC} \\ \hline
\multicolumn{1}{|l|}{\textbf{Mean-pooling}} & 0.8401              & 0.9786            \\ \hline
\multicolumn{1}{|l|}{\textbf{Max-pooling}}  & 0.9463              & 0.9879            \\ \hline
\multicolumn{1}{|l|}{\textbf{ABMIL}}        & 0.8656              & 0.9702            \\ \hline
\multicolumn{1}{|l|}{\textbf{PT-MTA}}       & 0.8299              & 0.9700            \\ \hline
\multicolumn{1}{|l|}{\textbf{MIL-RNN}}      & 0.9107              & -                 \\ \hline
\multicolumn{1}{|l|}{\textbf{DSMIL}}        & 0.8925              & 0.9841            \\ \hline
\multicolumn{1}{|l|}{\textbf{CLAM-SB}}      & 0.8818              & 0.9723            \\ \hline
\multicolumn{1}{|l|}{\textbf{CLAM-MB}}      & 0.9377              & 0.9799            \\ \hline
\multicolumn{1}{|l|}{\textbf{TransMIL}}     & 0.9603              & 0.9882            \\ \hline
\multicolumn{1}{|l|}{\textbf{\begin{tabular}[c]{@{}l@{}}TransMIL\\ (our folds)\end{tabular}}} & 0.957 & 0.985 \\ \hline
\multicolumn{1}{|l|}{\textbf{\begin{tabular}[c]{@{}l@{}}CCAN\\ (our folds)\end{tabular}}}     & 0.970 & 0.985 \\ \hline
\end{tabular}%
\end{table}
}

\begin{table}[]
\centering
\caption{Hyperparameters used to train the CCAN models.}
\label{tab:hyperparams}
\begin{tabular}{|l|c|}
\hline
\textbf{Batch size}       & 30               \\ \hline
\textbf{No. stages ($J$)} & 6                \\ \hline
\textbf{Optimizer}        & AdamW \cite{loshchilov_decoupled_2019}           \\ \hline
\textbf{Learning rate}    & 5e-6             \\ \hline
\textbf{Scheduler}        & Cosine annealing \cite{loshchilov_sgdr_2017} \\ \hline
\textbf{\begin{tabular}[c]{@{}l@{}}No. self-attention layers after each\\ cross attention\end{tabular}} & 2 \\ \hline
\textbf{Dropout rate $p_{do}$}           & 0.9              \\ \hline
\textbf{Dimensionality of latent tokens $D_l$}            & 512              \\ \hline
\textbf{Dimensionality of feature tokens $D_f$}            & 2048             \\ \hline
\textbf{No. latent tokens in the first stage $M$}              & 512              \\ \hline
\textbf{Compression factor $C$}              & 2                \\ \hline
\textbf{Maximal frequency $f_{max}$}        & 10               \\ \hline
\textbf{No. frequencies $I$}              & 6                \\ \hline
\end{tabular}%
\end{table}

\end{document}